# Lightweight True In-Pixel Encryption with FeFET-Enabled Pixel Design for Secure Imaging

Md Rahatul Islam Udoy, *Graduate Student Member, IEEE,* Diego Ferrer, *Graduate Student Member, IEEE,* Wantong Li, *Member, IEEE,* Kai Ni, *Member, IEEE,* Sumeet Kumar Gupta, *Senior Member, IEEE* and Ahmedullah Aziz, *Senior Member, IEEE*

*Abstract*— Ensuring end-to-end security in image sensors has become essential as visual data can be exposed through multiple stages of the imaging pipeline. Advanced protection requires encryption to occur before pixel values appear on any readout lines. This work introduces a secure pixel sensor (SecurePix), a compact CMOS-compatible pixel architecture that performs true in-pixel encryption using a symmetric key realized through programmable, non-volatile multidomain polarization states of a ferroelectric field-effect transistor. The pixel and array operations are designed and simulated in HSPICE, while a 45 nm CMOS process design kit is used for layout drawing. The resulting layout confirms a pixel pitch of 2.33 × 3.01 µm². Each pixel's non-volatile programming level defines its analog transfer characteristic, enabling the photodiode voltage to be converted into an encrypted analog output within the pixel. Full-image evaluation shows that ResNet-18 recognition accuracy drops from 99.29% to 9.58% on MNIST and from 91.33% to 6.98% on CIFAR-10 after encryption, indicating strong resistance to neural-network-based inference. Lookup-table-based inverse mapping enables recovery for authorized receivers using the same symmetric key. Based on HSPICE simulation, the SecurePix achieves a per-pixel programming power-delay product of 17 $\mu W \cdot \mu s$ and a per-pixel sensing power–delay product of 1.25 $\mu W \cdot \mu s$, demonstrating low-overhead hardware-level protection.

## I. INTRODUCTION

IMAGE sensors were originally developed for applications where security was not a design consideration. Early digital cameras, machine-vision sensors, and embedded imagers transmitted pixel data in clear form over electrical connections with no associated protection mechanism. This design philosophy persisted for years because imaging hardware was typically deployed in controlled environments, assumed to be physically secure, and not expected to handle sensitive information. As a result, most imaging pipelines were built around the idea that the sensor's role was simply to convert light into pixel values and pass those values forward without any form of confidentiality or integrity enforcement. As imaging systems migrated into open, networked, and user-centric environments, the consequences of transmitting unencrypted image data became increasingly severe [1], [2], [3]. When pixel data is placed on an unprotected transmission channel, any interception instantly reveals the full scene content. Even basic passive eavesdropping on wired links, system buses, ribbon cables, or unprotected wireless channels can expose private information, enable identity tracking, assist in unauthorized surveillance, or leak sensitive details [3], [4]. Once the risk of transmission-level attacks became clear, system designers introduced secure communication protocols and encrypted data paths to protect images as they leave the sensor module [5].

However, securing the transmission channel alone is not sufficient [6]. Even with encrypted external interfaces, the internal pathways that carry image data inside the system remain vulnerable [7]. The raw analog or lightly processed digital values generated by the sensor still traverse on-board buses, connector interfaces, and internal signal paths before reaching any encryption engine [8]. An adversary with access to these internal routes can tap the pixel stream directly and recover the image long before any transmission-level protection is applied [9]. This mode of attack does not require compromising the external link; it targets the wiring and nodes inside the device itself [7]. As adversaries have become more sophisticated, the attack boundary has continued to move inward toward the sensor. Hardware probing, bus tapping, fault injection, and machine-learning-based inference now allow attackers to extract information even from noisy or partial analog signals [10], [11]. In this context, the pixel array becomes the earliest point at which the image exists in physical form, and therefore the most critical location to protect. If the pixel outputs appear on shared readout lines in clear form, no amount of downstream encryption can prevent their exposure once the attacker is close enough to the hardware [12]. These limitations motivated the development of in-sensor processing, where some transformation or obfuscation is applied before data leaves the sensor chip. Yet most prior approaches implement their computation in shared peripheral circuits or column-level processing blocks [3], [4], [8]. While they strengthen security relative to unsecured transmission channels, they do not eliminate the fundamental exposure: the raw pixel output must still travel from the pixel to the peripheral block, and an attacker probing earlier in the pipeline can access it in plain form.

A stronger protection frontier is reached only when the

This work was supported in part by <funder name>

Md Rahatul Islam Udoy, Diego Ferrer, and Ahmedullah Aziz are with the Department of Electrical Engineering and Computer Science, University of Tennessee, Knoxville, TN 37996, USA (email: mudoy@vols.utk.edu; dferrer1@vols.utk.edu; aziz@utk.edu).

Wantong Li is with the Electrical and Computer Engineering Department, University of California, Riverside, CA 92521(email: wantong.li@ucr.edu).

Kai Ni is with the Electrical Engineering Department, University of Notre Dame, IN 46556 (email: kni@nd.edu).

Sumeet Kumar Gupta is with the Electrical and Computer Engineering Department, Purdue University, West Lafayette, IN 47907(email: guptask@purdue.edu).

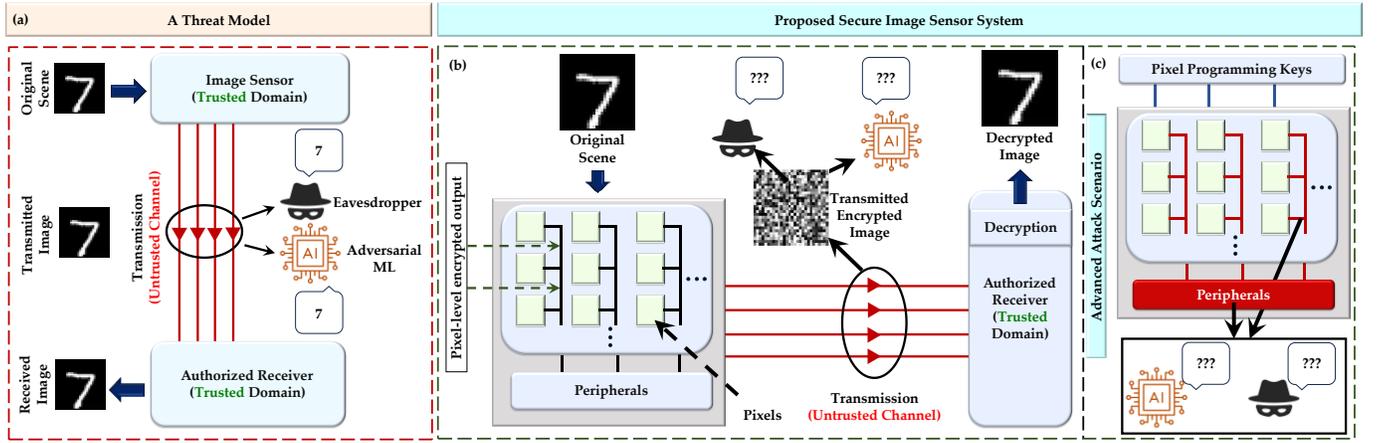

Fig. 1. (a) A threat model: The image sensor operates entirely in a trusted domain and produces an unencrypted pixel array. During transmission through an untrusted channel, both a passive eavesdropper and an ML-based adversary can directly access the unencrypted pixel stream, enabling unauthorized scene reconstruction or automated recognition. (b) Proposed secure image sensor system. Each pixel produces an encrypted output directly at the sensor array. The encrypted image is transmitted through the same untrusted channel; however, both eavesdroppers and ML-based attackers observe only visually unintelligible encrypted data. A trusted receiver, equipped with the correct per-pixel programming keys and inverse mapping, decrypts the image to recover the original scene. (c) Advanced threat model. A stronger adversary with partial access to sensor peripherals and individual pixel probing capability, but who still lacks the programming keys. As a result, even invasive or ML-assisted attacks remain unable to reconstruct the scene, demonstrating resilience of the proposed in-pixel encryption mechanism against both moderate and severe threat models.

security mechanism resides inside each pixel, ensuring that the signal leaving the pixel boundary is already encrypted. This true in-pixel processing prevents exposure of unencrypted data anywhere in the readout chain. Importantly, simply copying an existing encryption block from peripheral circuitry and embedding it inside each pixel is not a viable solution. Conventional encryption circuits occupy significant silicon area, rely on complex logic, and require routing resources that would drastically increase pixel pitch if replicated per pixel [13].

In this work, we introduce SecurePix, a compact and CMOS-compatible pixel architecture that embeds encryption directly inside each pixel through a programmable device state. Instead of producing raw pixel values, each pixel emits an encrypted signal whose form depends on a locally stored key. As a result, no unprotected image information ever leaves the pixel array, and all downstream circuitry, from on-chip wiring to peripheral readout blocks, sees only encrypted data. This design elevates the pixel itself into the first point of security, ensuring strong protection against attackers who may access internal sensor interfaces or shared signal paths.

## II. BACKGROUND & MOTIVATION.

This section outlines the key concepts needed for SecurePix. We first illustrate how conventional imaging pipelines expose unprotected data at multiple points and how stronger threat models progressively push the security boundary toward the pixel array. We then review the standard 3T pixel operation and the FeFET device behavior that enables programmable analog modulation.

### A. Threat Model

Security vulnerabilities manifest at different points in the imaging pipeline depending on where pixel values become exposed. Fig. 1(a) shows a baseline threat model in which the sensor chip is trusted, but the unencrypted pixel stream travels through an untrusted transmission channel. If the pixel data leaves the chip in clear form, this model cannot be defended because an eavesdropper or ML-based adversary can directly access the raw scene [14], [15]. To address this, we introduce SecurePix, an architecture that performs encryption inside each pixel. If the pixel array output is already encrypted before it reaches any shared line, as in Fig. 1(b), then transmission-level attacks no longer reveal meaningful information because only encrypted intensities propagate through the channel. A stronger adversary may attempt to probe internal sensor wiring or peripheral circuits, as illustrated in Fig. 1(c); however, if unencrypted photodiode values never appear on these internal nodes, then this more advanced threat model is also neutralized. We propose that SecurePix can withstand both of these moderate and severe attack scenarios by ensuring that encryption is enforced at the pixel boundary.

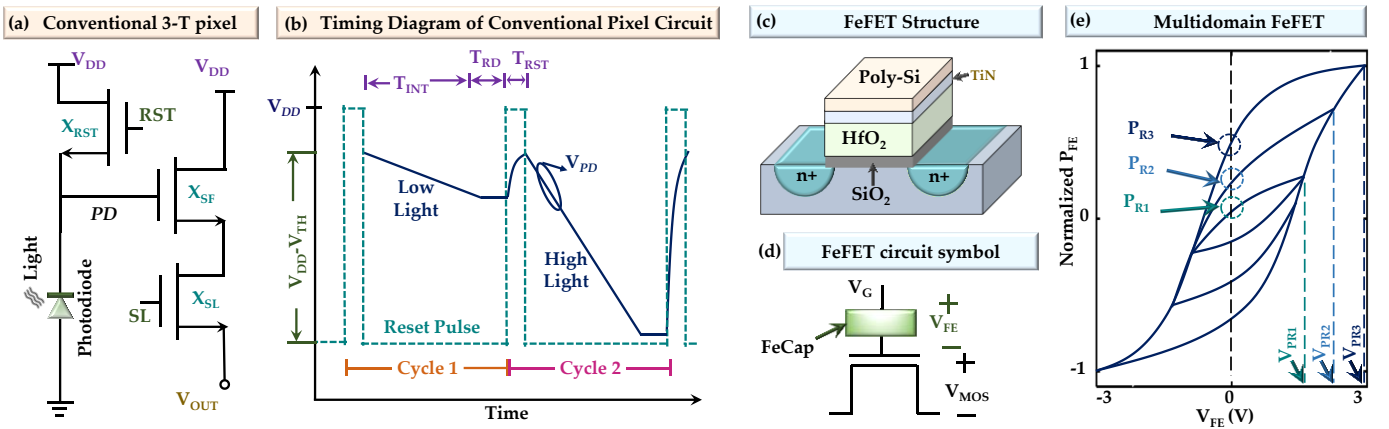

Fig. 2. (a) Schematic of a standard 3-transistor (3-T) active pixel circuit where, $X_{RST}$, $X_{SF}$, and $X_{SL}$ denote the reset, source-follower, and selector transistors, respectively. (b) Timing diagram of the 3-T pixel operation illustrating the integration ($T_{INT}$), readout ($T_{RD}$), and reset ($T_{RST}$) phases. (c) Structure of an HfO2-based ferroelectric FET (FeFET). (d) Circuit-level symbol of the FeFET. (e) Multidomain polarization characteristics of the FeFET, highlighting multiple remnant polarization ($P_R$) states for different programming voltage levels ($V_{PR}$).

### B. Pixel Circuit

The main building block of an image sensor chip is the two-dimensional (2D) pixel array [16]. Each pixel of the array commonly uses a 3-transistor (3T) pixel circuit (Fig. 2(a)), which consists of a reverse-biased photodiode and three transistors: $X_{RST}$ for reset, $X_{SF}$ as the source follower, and $X_{SL}$ as the selector transistor [17]. The overall operation is shown in Fig. 2(b). The circuit function begins with the reset phase. When $X_{RST}$ is turned on, the photodiode node ($V_{PD}$) is charged to roughly $V_{DD}$ minus the threshold voltage of the reset transistor. If a PMOS device is used instead, the PD node can be charged fully to $V_{DD}$ [18]. After reset, $X_{RST}$ is turned off and the light integration phase starts ($T_{INT}$). During this period, the photodiode generates photocurrent in response to incident light, which causes $V_{PD}$ to drop gradually over time. The rate of this drop follows the simple relation $dV_{PD}/dt = I_{PD}/C_{PD}$. Under brighter illumination, the photocurrent is larger, so $V_{PD}$ decreases more quickly, as seen in cycle 2 of Fig. 2(b). The source-follower transistor $X_{SF}$ passes the PD node voltage to the readout node while maintaining the stored charge, and $X_{SL}$ selects the pixel during readout phase ($T_{RD}$). The pixel output is referred to as $V_{OUT}$.

### C. FeFET

A Ferroelectric Field-Effect Transistor (FeFET) is a non-volatile transistor that embeds a ferroelectric capacitor (FeCap) into the gate stack of a standard MOSFET [19]. An example structure using $HfO_2$ as the ferroelectric layer is shown in Fig. 2(c). The defining property of a ferroelectric material is its ability to maintain a spontaneous polarization even after the external electric field is removed; this is referred to as the remnant polarization ($P_R$) [20]. When the device is programmed with an appropriate gate-voltage pulse, this remnant polarization determines the channel conductivity [21].

FeFETs are compatible with conventional CMOS processing, making them well suited for integration in emerging computing systems [22]. Traditional FeFETs operate using two polarization states, but recent material improvements have enabled multidomain behavior within the ferroelectric film [23]. In these multidomain FeFETs, the polarization is spread across multiple nanoscale domains that can switch independently. As a result, the device can hold several stable intermediate polarization levels, giving rise to a quasi-analog modulation of channel conductance [23]. Fig. 2(e) illustrates this multi-level behavior, where different programming voltages ($V_{PR1}$, $V_{PR2}$, and $V_{PR3}$) produce different remnant polarization states ($P_{R1}$, $P_{R2}$, and $P_{R3}$). In this work, we utilize this multi-level polarization property of an FeFET to design our pixel circuit.

### III. SECUREPIX SENSOR

The proposed SecurePix framework introduces encryption at the earliest physically accessible point in the imaging pipeline: the pixel itself. Instead of transmitting the photodiode voltage in plaintext, the proposed pixel transforms the signal into an encrypted analog quantity.

### A. Circuit-Level Design & Operation

The proposed pixel circuit configuration is shown in Fig. 3(a). Because we choose 1V as the $V_{DD}$, a PMOS device is used for the reset transistor ($X_{RST}$). An NMOS reset transistor would introduce a threshold-related voltage drop that significantly reduces the available signal swing in low-voltage operation. The gate of this $X_{RST}$ is controlled by the signal RST. The $X_P$ transistor is used to sense the PD node voltage. An n-type FeFET ($X_{FE}$) is connected to the source terminal of $X_P$. The $X_{FE}$ gate is controlled by the signal $V_G$, which is the applied programming voltage. We use the discharge transistor $X_{DIS}$ to connect the source terminal of the FeFET to ground during programming. The gate of this transistor is controlled by the signal Dis. The selector transistor $X_{SL}$ performs the standard pixel selection operation, and this is controlled by the gate signal SL. The output of the pixel is the current $I_{OUT}$, which can be converted to a voltage using a transimpedance amplifier.

We utilize the multidomain ferroelectric capacitor (FeCap) model from [20], which has been rigorously calibrated to experimental ferroelectric measurements and verified to reproduce the essential switching and polarization characteristics of FeFET devices. As illustrated in Fig. 3(b), both the FeCap and MOSFET models incorporate an auxiliary terminal to facilitate charge transfer (Q). In practice, this extra terminal is only required for the FeFET ($X_{FE}$), where the ferroelectric polarization charge must be coupled to the MOSFET channel current to accurately represent multidomain ferroelectric behavior. In Fig. 3 (b), $V_{FE}$ is the voltage across the FeCap, $V_{MOS}$ is the voltage across the internal gate to source voltage of the MOSFET. All remaining transistors in the circuit are implemented using conventional four-terminal MOSFET models.

We simulate the proposed pixel circuit in HSPICE, an industry simulation tool [24]. The transistors are modeled using 45-nm high-performance CMOS parameters from the Predictive Technology Model [25], which has been widely validated and is regarded as a reliable representation of experimentally measured device behavior. The overall pixel operation consists of two major phases: the programming phase and the imaging phase. During the programming phase, the discharge transistor $X_{DIS}$ is turned on, as illustrated in Fig. 3(c).

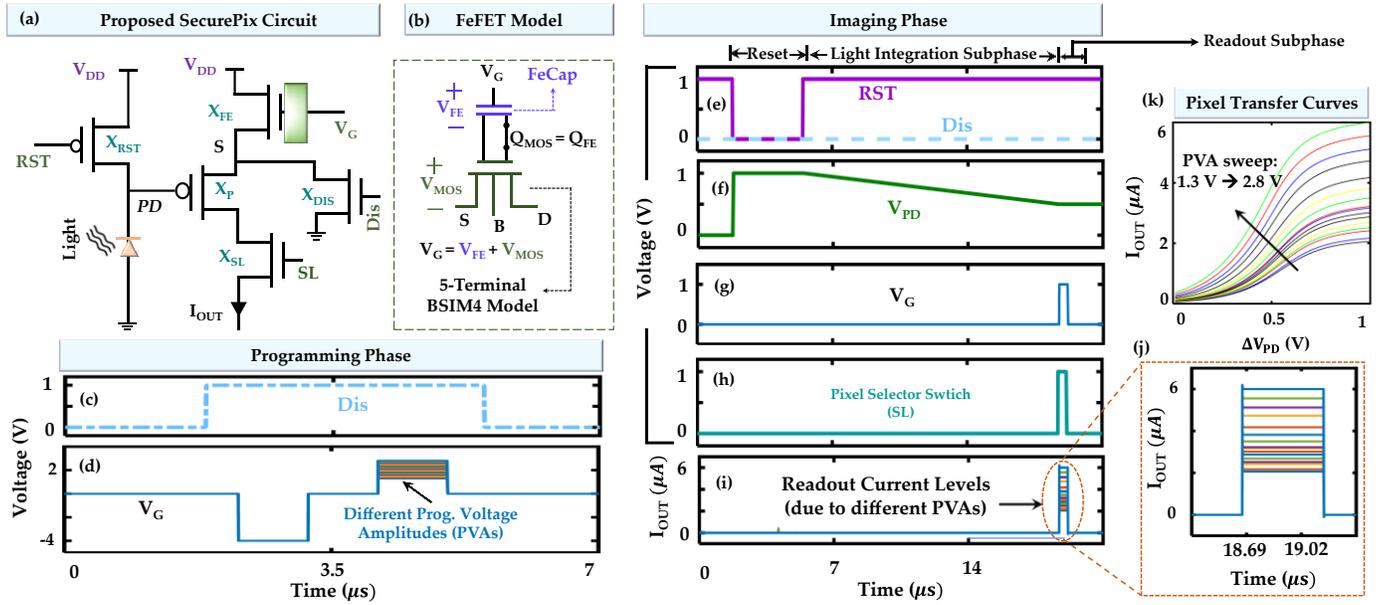

**Fig. 3. (a)** Schematic of the proposed SecurePix circuit, where the FeFET ($X_{FE}$) modulates the pixel output based on its programmed polarization state. The discharge transistor ($X_{DIS}$) connects the FeFET source to ground during the programming phase, so that the FeFET can utilize the full programming gate voltage ($V_G$). The $X_P$ transistor is used to transfer the PD node voltage to the readout branch. The remaining transistors are the reset ($X_{RST}$) and selector ($X_{SL}$) transistors. **(b)** FeFET compact model representation using a modified 5-terminal MOSFET and an HfO$_2$-based ferroelectric capacitor (FeCap). The additional terminal is used to pass the value of charge. **(c)** $X_{DIS}$ gate control signal (Dis) is activated during the programming phase to connect the $X_{FE}$ source to the ground. **(d)** Programming waveform applied to the FeFET gate ($V_G$), demonstrating multiple programming voltage amplitudes (PVAs) used to create distinct polarization states. **(e-i)** Imaging-phase operation of the proposed pixel, which consists of the reset, light-integration, and readout phases. **(e)** The PMOS reset transistor is turned on during reset phase. The Dis signal remains inactive throughout the imaging phase. **(f)** Photodiode voltage ($V_{PD}$) discharge trajectory under light integration. **(g)** FeFET and **(h)** selector transistors are turned on during the readout phase. **(i)** For a single level of light exposure (i.e., a single $V_{PD}$ value at readout), multiple output current levels are produced, each corresponding to a different programmed voltage amplitude (PVA) of the FeFET. **(j)** Zoomed-in view of the multi-level readout currents shown in **(i)**. **(k)** Pixel transfer curves generated by sweeping the PVA from 1.3 V to 2.8 V, demonstrating programmable, multi-level analog modulation of the pixel response.

Enabling $X_{DIS}$ forces the source terminal of the FeFET to be pulled close to ground potential. This step is essential because the effective programming voltage of an FeFET is determined by the difference between the applied gate voltage $V_G$ and the source voltage $V_S$. By ensuring that $V_S$ is held at zero during programming, the FeFET experiences the full applied programming amplitude, which maximizes the polarization shift that can be induced in the ferroelectric layer. As shown in Fig. 3(d), before setting the FeFET to the desired state, a negative gate pulse is first applied to drive the device toward the outer branch of its hysteresis loop. This step ensures that the initial ferroelectric domain configuration is well defined and consistent across programming cycles. Once the negative pulse has established this baseline state, a positive programming pulse is applied. The amplitude of this positive pulse, referred to as the programming-voltage amplitude (PVA), determines the amount of remnant polarization that will be stored in the device. Larger PVAs switch a greater fraction of the ferroelectric domains, resulting in higher conductance states, while smaller PVAs produce intermediate levels of polarization. To fully characterize the programmability of the pixel, we apply a range of different PVAs to $X_{FE}$ and evaluate the resulting conductance levels during subsequent imaging operation. These effects, and their impact on the pixel output current, are examined in detail in the imaging-phase discussion.

The imaging phase is divided into three subphases: reset, light integration, and readout. The sequence begins with the reset subphase, shown in Fig. 3(e). In this step, the reset transistor $X_{RST}$ is turned on, which charges the photodiode node to $V_{DD}$. This action establishes the initial photodiode voltage $V_{PD}$ at the start of the frame, as depicted in Fig. 3(f). Once the reset signal is released, the circuit transitions into the light integration subphase. During this period, incident photons generate a photocurrent within the photodiode, causing $V_{PD}$ to decrease over time. The resulting voltage drop $\Delta V_{PD}(=V_{DD}-V_{PD})$, is directly proportional to the incident light intensity and represents the amount of accumulated charge. Following the light integration subphase, the readout subphase begins. As illustrated in Fig. 3(g) and Fig. 3(h), the selector transistor $X_{SL}$ and the FeFET $X_{FE}$ are enabled during this interval. The output current is determined by both $\Delta V_{PD}$ and the programmed state of the FeFET. The polarization level created during the programming phase determines the conductance of $X_{FE}$, so different PVAs produce distinct output current levels for the exact same amount of light. This behavior is shown in Fig. 3(i) and Fig. 3(j), where a fixed photodiode voltage results in multiple output current levels depending on the selected PVA.

Finally, to construct a compact pixel model for array-level simulation, we extract the input-to-output characteristics of the circuit from HSPICE. This relationship, shown in Fig. 3(k), maps the photodiode voltage drop that occurs during illumination to the corresponding pixel output current.

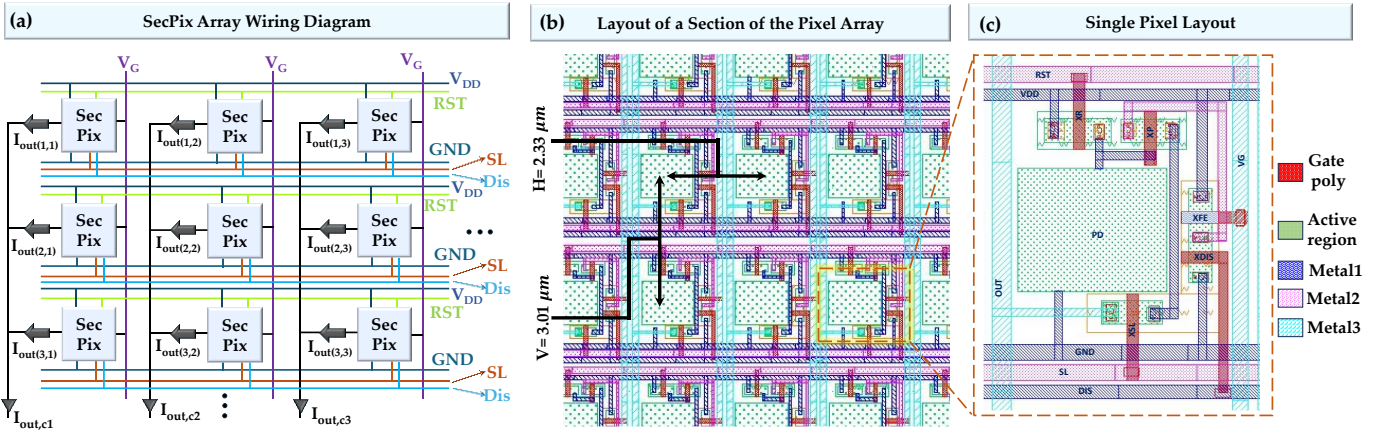

**Fig. 4. (a)** Array-level wiring diagram. All pixels in the same column share the FeFET gate-control line ($V_G$) and the pixel output metal lines. All pixels in the same row share the $V_{DD}$, RST, GND, SL and Dis metal lines. **(b)** Layout of a section of the SecurePix array, implemented in Cadence Virtuoso using a 45-nm CMOS PDK. The figure shows the tiled arrangement of individual pixels along with the horizontal (≈2.33 μm) and vertical (≈3.01 μm) array pitch. **(c)** Detailed layout of a single SecurePix pixel.

### B. Array-Level Design & Operation

The pixel circuit described in the previous section is embedded within a two-dimensional array, and its behavior is strongly influenced by the organization of the shared row and column signals. To understand how the FeFET states are programmed and how the encrypted pixel outputs propagate through the array, it is necessary to examine the global wiring structure that interconnects the pixels. Fig. 4(a) presents the array-level wiring diagram, which highlights the set of signals that are shared along each row and column. Each pixel contains the circuit described in Section III (A). All pixels within the same column share a common $V_G$ line, which distributes the programming voltage to every FeFET in that column. Along each row, the pixels share the signals RST, SL, $V_{DD}$, GND, and Dis. These row-wise lines allow the control logic to independently activate a specific row for programming or imaging by enabling the appropriate combination of reset, selector, and discharge signals. The output currents from all pixels in a column flow into dedicated column readout branches labeled $I_{out,cn}$ (where n is the column number). These lines route the encrypted pixel currents to downstream conversion circuitry. This wiring configuration forms the foundation for both the row-wise programming protocol and the encrypted imaging operation described later in this section.

To ensure that the proposed pixel and array architecture can be physically realized within standard design rules, we implemented the layout using a 45 nm process design kit (PDK). Fig. 4(b) and (c) show the layout implementation of the proposed SecPix architecture. The layouts were created in Cadence Virtuoso and were verified through Design Rule Checking (DRC). Fig. 4(b) illustrates a section of the pixel array, where the repeated pixel structures are clearly visible. This view highlights how the row-wise and column-wise routing described earlier is translated into a layout implementation. Fig. 4(c) shows the layout of a single pixel. It is important to note that, to represent the FeFET in the layout, we use a regular NMOS device whose gate area is sized to accommodate the equivalent ferroelectric capacitor. The various dimensions extracted from the layout are summarized in Table I. Pixel pitch was measured based on the center-to-center separation between adjacent pixels in both horizontal and vertical directions. The active photodiode size is consistent with sizes reported in prior works [26], [27], [28]. The fill factor (FF) reported in Table I corresponds to this specific layout and is not a fundamental limitation of the SecurePix architecture. The fill factor can be increased as needed by allocating a larger photodiode area, and even 100% FF is achievable in a stacked sensor implementation.

#### 1) Programming, Encryption & Imaging

We now describe the programming protocol in a step-by-step manner. Programming the FeFETs in the SecurePix array is performed in a row-wise manner, enabled by the shared $V_G$ lines along columns and the SL and Dis lines along rows. This approach allows each pixel to receive its intended programming-voltage amplitude. The programming process begins with an initialization stage, illustrated in Fig. 5(a). During this stage, each pixel is scheduled to receive a target programming level, represented as L1, L2, L3, and so on, corresponding to different FeFET polarization states. These levels serve as the programming keys for image encryption. Fig. 5(b) shows the programming of the first row. The desired programming-voltage levels for this row are applied simultaneously to the $V_G$ lines. To activate the first row, the SL1 and Dis1 signals are asserted, which turn on both the selector and discharge transistors in that row. When $X_{DIS}$ is enabled, the source terminal of each FeFET in the row is pulled close to ground. This ensures that the effective programming voltage, equal to $V_{GS}$, reaches its maximum value, enabling accurate polarization programming of the FeFETs in the selected row. The programming then proceeds to the next row, shown in Fig. 5(c). The $V_G$ lines are updated with the programming levels intended for the second row, and the SL2 and Dis2 signals are activated to select this row. While the second row undergoes programming, all previously programmed rows remain protected because their SL and Dis signals stay inactive. This keeps their FeFET source voltages elevated and reduces the effective programming voltage to a level too small to modify their stored polarization states. As a result, earlier rows retain their programmed conductance states. In this way, the entire array becomes programmed, and the resulting set of programming levels acts as the key for the SecurePix system.

TABLE I
Layout Dimensions

| Dimension | Value | Dimension | Value |
|---|---|---|---|
| Technology Node | 45 nm | Photodiode Width | 1.3825 μm |
| Horizontal Pixel Pitch | 2.33 μm | Photodiode Height | 1.165 μm |
| Vertical Pixel Pitch | 3.01 μm | Photodiode Area | 1.611 μm² |
| Pixel Area | 7.0133 μm² | Fill Factor | 23% |

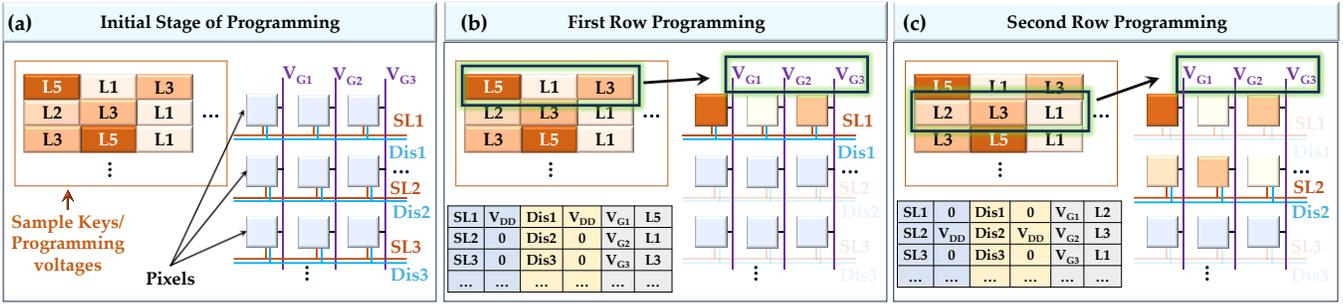

**Fig. 5.** Row-by-row FeFET programming procedure in the SecurePix array. **(a)** Initial stage of programming, where each pixel is assigned a target programming-voltage level (L1, L2, L3 …), corresponding to the desired FeFET polarization state. The $V_G$ lines distribute the programming voltages, while the Dis and SL lines determine which row is actively programmed. **(b)** First-row programming. Sample programming voltages (L5, L1, L3,…) for the first row are loaded in the $V_G$ lines. The SL1 and Dis1 lines are activated, turning on both the selector and discharge transistors in the first row. This pulls the FeFET source nodes close to ground, ensuring a large effective programming voltage ($V_G$-$V_S$) and enabling proper polarization programming. **(c)** Second-row programming. Sample programming voltages (L2, L3, L1,…) for the second row are loaded in the $V_G$ lines. The SL2 and Dis2 lines are activated and the second row become programmed. Previously programmed rows remain protected because their discharge and selector transistors are turned off, resulting in insufficient effective programming voltages ($V_G$-$V_S$) to alter their FeFET states.

Once the programming phase is completed, the SecurePix array operates as a standard image sensor during the imaging phase. The programming step is required only once, since the FeFET stores its programmed polarization state nonvolatilely. The imaging process can therefore proceed continuously without reprogramming unless a new key is desired. During imaging, all Dis lines remain deactivated, which isolates the FeFET source nodes from ground and ensures that no unintentional polarization change occurs. The reset, integration, and readout subphases follow the conventional timing of a three-transistor pixel. The photodiode voltage decreases according to the incident illumination, and the readout current reflects both this photodiode voltage and the conductance state of the FeFET that was programmed earlier. Consequently, each pixel exhibits its own distinct input-output transfer curve determined by its programmed FeFET state. The output from each pixel is encrypted, so even if an attacker probes the column readout lines or any downstream processing peripherals, the attacker observes only encrypted values. These per-pixel transfer characteristics introduce spatially varying modulation across the image, forming the basis of pixel-level encryption. Since the FeFET retains its polarization even after power is removed, the encryption key remains stored across the array without requiring any refresh or energy consumption. HSPICE simulations show that SecurePix provides low-overhead hardware security, with a per-pixel programming power-delay product (PDP) of 17 μW.μs and a per-pixel sensing PDP of 1.25 μW.μs.

*2) Test Setup & Results*

We now describe how we evaluate the impact of the proposed pixel design at the full-image level. To mimic pixel-to-pixel process variation and noise introduced by the FeFET, we perform a Monte Carlo analysis with 10,000 samples, where key device parameters are perturbed according to Gaussian distributions. In particular, the FeCap thickness is modeled with a nominal value of 4 nm and a standard deviation corresponding to ±9% variation, and the threshold voltages of $X_{FE}$ and $X_P$ are varied within ±150 mV. For each pixel in the array, we assign a programming level (key) and associate it with a specific Monte Carlo sample, so that every pixel experiences a distinct combination of FeFET state and process variation.

To test the effect on complete images, we first convert the pixel values of a given input image into corresponding photodiode voltages. These voltages are then applied to the pixel array in HSPICE, and the resulting outputs from all pixels are collected. The simulated analog outputs are converted into 8-bit digital values and reconstructed into images using MATLAB. Fig. 6(a)-(d) present several examples, where the first image in each set shows the original input and the second image shows the corresponding encrypted version produced by the SecurePix array under influence of programming keys. Table II summarizes several statistical metrics extracted from encrypted versions of two real world example test images: a vehicle number-plate image and the Barbara test image (Fig. 6(a)-(b)) [29]. For each case, the table reports the horizontal, vertical, and diagonal correlation coefficients of both the original and encrypted images, along with the PSNR computed between the encrypted image and its corresponding original version. As seen in the table, the original images exhibit very strong pixel-to-pixel correlations, as expected for natural scenes. After encryption through the SecurePix array, these correlations drop sharply to values near zero, indicating that the encrypted outputs retain almost no spatial redundancy. The PSNR values between encrypted and original images further confirm that the encrypted images diverge substantially from the originals, consistent with strong visual obfuscation.

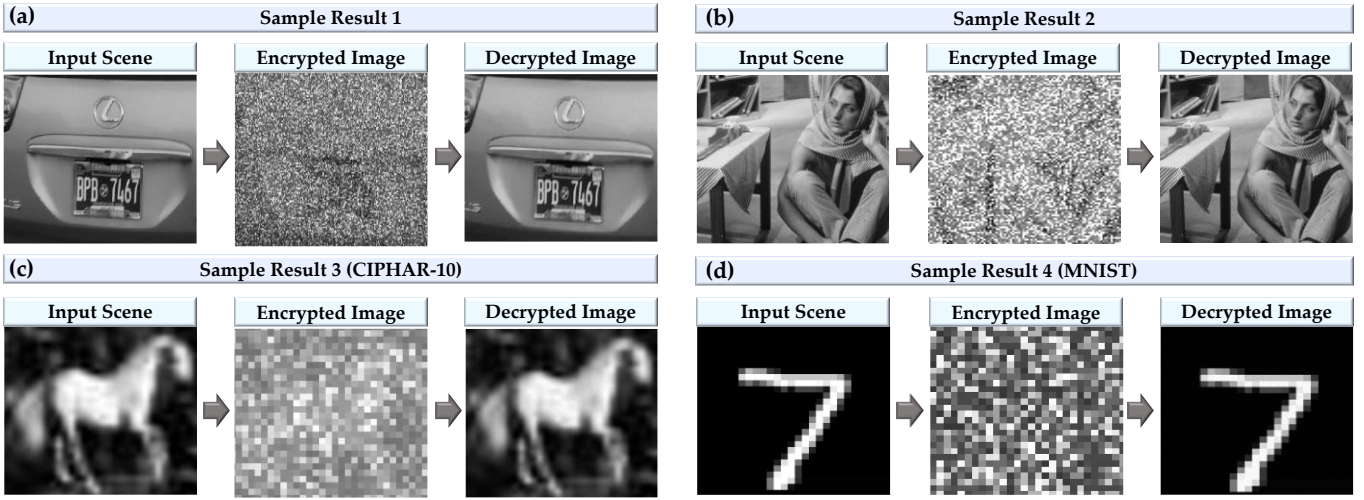

**Fig. 6.** Sample encryption and decryption results produced by the proposed SecurePix system. **(a–d)** Each subfigure shows the input scene, the corresponding encrypted image generated by in-pixel FeFET modulation, and the decrypted reconstruction at the authorized receiver. **(a)** Real-world license plate example. **(b)** The standard Barbara test image. **(c)** CIFAR-10 sample. **(d)** MNIST digit sample. In all cases, the encrypted images become visually noise-like and unintelligible, while the decrypted outputs faithfully recover the original content.

TABLE II
Results for Sample Real-World Test Images

| Test Image | Horizontal Corr. | Vertical Corr. | Diagonal Corr. | PSNR (Enc vs. Orig) [dB] | PSNR (Recovery) [dB] |
|---|---|---|---|---|---|
| Number Plate (Fig. 6(a)) [Original] | 0.9840 | 0.9709 | 0.9529 | 12.3486 | 47.70 |
| Number Plate (Fig. 6(a)) [Encrypted] | 0.0959 | 0.0759 | 0.0680 | | |
| Barbara (Fig. 6(b)) [Original] | 0.8948 | 0.9585 | 0.8824 | 12.0533 | 46.11 |
| Barbara (Fig. 6(b)) [Encrypted] | 0.0884 | 0.0925 | 0.0851 | | |

To further evaluate the robustness of SecurePix against machine-learning-based attacks, we performed an image-classification test using a ResNet-18 neural network [30]. In this experiment, the classifier is treated as an adversarial model attempting to recognize the encrypted images. For CIFAR-10 [31], the ResNet-18 network was first trained only on the unencrypted training images following standard supervised-learning procedures. After training, we encrypted 10,000 CIFAR-10 test images using SecurePix and fed these encrypted images to the trained classifier. As shown in Table III, while the network achieves 91.33% accuracy on clean (unencrypted) CIFAR-10 test images, its accuracy drops dramatically to 6.98% on the encrypted images, which is close to random guessing for a 10-class dataset. A similar evaluation was performed for MNIST [32]: ResNet-18 trained on unencrypted MNIST images reached 99.29% accuracy on clean test data, but only 9.58% accuracy on encrypted MNIST images. These results demonstrate that the encrypted outputs of SecurePix severely degrade the recognizability of the images to modern deep neural networks, indicating that the proposed in-pixel encryption provides strong resistance against ML-based inference attacks.

TABLE III
ResNet18 Recognition Accuracy on Unencrypted vs. Encrypted Images

| Unencrypted MNIST Images | 99.29% | Unencrypted CIFAR-10 Images | 91.33% |
|---|---|---|---|
| Encrypted MNIST Images | 9.58% | Encrypted CIFAR-10 Images | 6.98% |

The number of available programming levels plays a crucial role in determining both the encryption strength and the recovery robustness of SecurePix. If each pixel can be programmed into one of $L$ distinct FeFET states, the per-pixel key entropy scales as $\log_2 L$ bits. $L=16$ provides 4 bits/pixel, while reducing to $L=8$ lowers this to 3 bits/pixel, directly shrinking the key space and reducing the diversity of encrypted outputs. At the device level, prior work on multilevel FeFETs shows that the achievable $L$ is fundamentally limited by the ferroelectric memory window and the statistical spread of intermediate programmed states [33]. These depend on ferroelectric thickness, gate-stack design, and programming conditions, with experimental studies generally reporting 4-8 reliably separable levels under realistic variability constraints [34], [35]. Reducing $L$ generally improves the separation between programmed FeFET states, making recovery more robust under device variation, while increasing $L$ strengthens encryption but demands tighter control over FeFET polarization levels. Consequently, the choice of $L$ reflects a practical trade-off between encryption complexity and reliable image reconstruction.

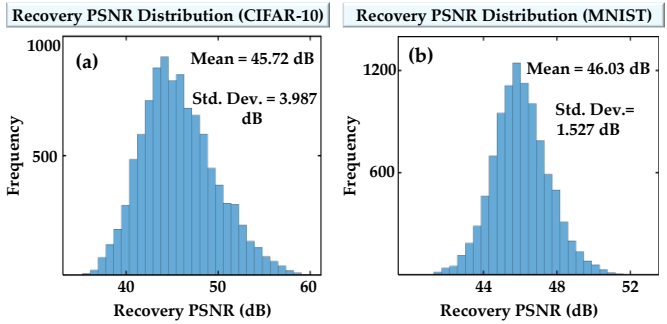

Fig. 7. Recovery PSNR distribution across large-scale test sets for the proposed SecurePix system. **(a)** CIFAR-10 recovery PSNR distribution evaluated over 10,000 images. **(b)** MNIST recovery PSNR distribution evaluated over 10,000 images. Both histograms demonstrate consistently high reconstruction fidelity across the full datasets.

### C. Decryption

While the previous subsections focused on how the SecurePix array performs pixel-level encryption during imaging, it is equally important to explain how an authorized receiver can correctly reconstruct the original image. The decryption procedure relies on the fact that the receiver has full knowledge of the programming keys that were used to configure the FeFETs. Since each programming level corresponds to a distinct input-output transfer curve, knowing the key for each pixel is equivalent to knowing which curve that pixel followed during the imaging process. Prior to operation, these transfer characteristics are extracted from circuit simulations and stored as lookup tables. During decryption, the encrypted pixel values are individually mapped back through the corresponding lookup tables to recover estimates of the original photodiode voltages. This pixel-wise inverse mapping produces the reconstructed image, examples of which are shown in the third column of Fig. 6 for all four sample scenes. The last column of Table II reports the PSNR between the original and decrypted images for the number-plate and Barbara examples, demonstrating high recovery fidelity. To evaluate the recovery accuracy across larger datasets, we computed the same recovery PSNR metric for 10,000 CIFAR-10 images and 10,000 MNIST images. The statistical distributions of these results are presented in Fig. 7, showing consistently strong reconstruction quality across both datasets.

## IV. DISTINCTIONS FROM EXISTING IN-SENSOR ENCRYPTION METHODS

Several prior works have proposed securing image sensors through in-sensor or near-sensor cryptographic techniques [3], [4], [8]. Although all of these approaches enhance security relative to conventional image sensors, their mechanisms differ fundamentally from the proposed SecurePix architecture. In the prior designs, encryption or obfuscation is performed in peripheral compute blocks that operate after the pixel photoconversion stage. Consequently, the pixel outputs themselves remain unencrypted as they propagate through shared readout paths, leaving portions of the analog front-end exposed to adversarial observation or manipulation. In contrast, SecurePix introduces encryption directly inside the pixel circuit. Because the encryption occurs before any shared column circuitry, the attack surface is significantly reduced, and adversaries probing the column buses or downstream analog blocks observe only encrypted signals. Given the architectural mismatch between SecurePix and prior systems, particularly the absence of in-pixel modification in earlier works, traditional quantitative tables comparing transistor count, pitch, or analog performance would not be meaningful or fair. Moreover, metrics such as NPCR (Number of Pixel Change Rate) and UACI (Unified Average Changing Intensity) are not applicable here, because SecurePix encrypts each pixel independently without cross-pixel diffusion, which these metrics inherently assume. Instead, Table IV provides comparison that highlights the fundamental distinctions across representative prior techniques. These distinctions guide how different approaches can be combined rather than positioned as direct competitors. These distinctions clarify that the methods target different points in the imaging pipeline and therefore serve complementary roles. SecurePix provides protection at the pixel level before any shared circuitry, while prior techniques operate later in the readout or processing chain. When used together, these layers can reinforce each other, creating a more robust and comprehensive secure-imaging system than any single method alone.

TABLE IV
Comparison Table

| Work | Where encryption occurs | Technology Style | Exposure of unencrypted data |
|---|---|---|---|
| This work | Inside pixel | CMOS (45 nm) | Never exposed (encrypted in pixel) |
| [3] | Peripheral arithmetic blocks | CMOS | Exposed before chaotic mixing |
| [4] | Peripheral signal processing | CMOS | Exposed before chaotic mixing |
| [8] | Memristor array processing (post-sensing) | Optoelectronic / memristive | Remain exposed before memristor encryption |

## V. CONCLUSION

SecurePix demonstrates that embedding programmable, non-volatile ferroelectric states inside each pixel enables true in-pixel encryption that protects pixel information before it enters the readout columns. The HSPICE-based pixel and array simulations show that the programmed transfer characteristics suppress spatial correlation, disrupt machine-learning recognizability, and still allow accurate reconstruction for authorized receivers through lookup-table inversion. Because SecurePix operates at the pixel boundary, it addresses a security point not covered by most of the prior in-sensor or near-sensor encryption schemes, which act only after photoconversion. These distinctions indicate that SecurePix and existing approaches serve different roles in the imaging pipeline: SecurePix secures the pixel output itself, while prior methods protect later stages. Combined, they can form a more robust and comprehensive secure-imaging system.